\documentclass{article} % For LaTeX2e
\usepackage[final]{colm2026_conference}
\usepackage{microtype}
\usepackage{hyperref}
\usepackage{url}
\usepackage{booktabs}
\usepackage{graphicx}
\usepackage{subcaption}
\usepackage{multirow}
\usepackage{wrapfig}
\usepackage{tikz}
\usepackage{amsmath}
\usepackage{amssymb}
\usepackage{lineno}

\definecolor{darkblue}{rgb}{0, 0, 0.5}
\hypersetup{colorlinks=true, citecolor=darkblue, linkcolor=darkblue, urlcolor=darkblue}

\title{Expert-Choice Routing Enables Adaptive Computation \\ in Diffusion Language Models}

\author{
\textbf{Shuibai Zhang\textsuperscript{*,1}} \quad
\textbf{Caspian Zhuang\textsuperscript{*,2}} \quad
\textbf{Chihan Cui\textsuperscript{*,1}} \quad
\textbf{Zhihan Yang\textsuperscript{3}} \\
\textbf{Fred Zhangzhi Peng\textsuperscript{4}} \quad
\textbf{Yanxin Zhang\textsuperscript{1,5}} \quad
\textbf{Haoyue Bai\textsuperscript{1}} \quad
\textbf{Zack Jia\textsuperscript{2}} \\
\textbf{Yang Zhou\textsuperscript{6}} \quad
\textbf{Guanhua Chen\textsuperscript{$\dagger$,7}} \quad
\textbf{Ming Liu\textsuperscript{$\dagger$,1}} \\[6pt]
\textsuperscript{1}University of Wisconsin-Madison \quad
\textsuperscript{2}Scitix \quad
\textsuperscript{3}Cornell University \quad
\textsuperscript{4}Duke University \\
\textsuperscript{5}NVIDIA \quad
\textsuperscript{6}UC Davis \quad
\textsuperscript{7}Southern University of Science and Technology \\[4pt]
{\small \textsuperscript{*}Equal contribution \quad \textsuperscript{$\dagger$}Equal advising}
}
\begin{document}

\ifcolmsubmission
\linenumbers
\fi

\maketitle

\begin{abstract}
Diffusion language models (DLMs) enable parallel, non-autoregressive text generation, yet existing DLM mixture-of-experts (MoE) models inherit token-choice (TC) routing from autoregressive systems, leading to load imbalance and rigid computation allocation.
We show that \textbf{expert-choice (EC) routing is a better fit for DLMs}: it provides deterministic load balancing by design, yielding higher throughput and faster wall-clock convergence than TC.
Building on the property that EC capacity is externally controllable, we introduce \emph{timestep-dependent expert capacity}, which varies expert allocation according to the denoising step. We find that \textbf{allocating more capacity to low-mask-ratio steps consistently achieves the best performance under matched FLOPs}, and provide a mechanistic explanation: tokens in low-mask-ratio contexts exhibit an order-of-magnitude higher learning efficiency, so concentrating compute on these steps yields the largest marginal return.
Finally, we show that \textbf{existing pretrained TC DLMs can be retrofitted to EC} by replacing only the router, achieving faster convergence and improved accuracy across diverse downstream tasks.
Together, these results establish EC routing as a superior paradigm for DLM MoE models and demonstrate that computation in DLMs can be treated as an adaptive policy rather than a fixed architectural constant. Code is available at \url{https://github.com/zhangshuibai/EC-DLM}.
\end{abstract}

%================================================================
\section{Introduction}
%================================================================

\begin{wrapfigure}{r}{0.48\textwidth}
  \centering
  \vspace{-12pt}
  \includegraphics[width=0.46\textwidth]{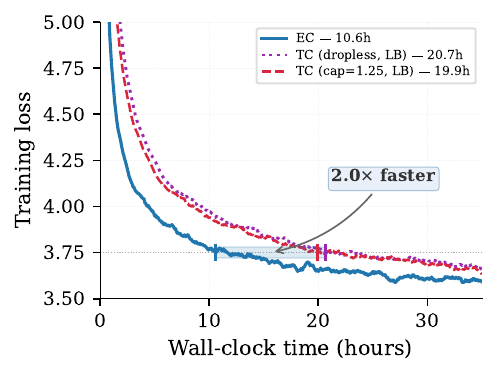}
  \caption{Training loss vs.\ wall-clock time. EC reaches loss 3.75 in 10.6h, $2.0\times$ faster in wall-clock time than TC (20.7h).}
  \label{fig:loss_wallclock}
  \vspace{-10pt}
\end{wrapfigure}

Diffusion language models (DLMs) have emerged as a promising alternative to autoregressive generation, enabling parallel decoding through iterative denoising~\citep{nie2025llada,ye2025dream,arriola2025block,liu2025tidar,liu2025wedlm}. Scaling language models to hundreds of billions of parameters increasingly relies on mixture-of-experts (MoE) architectures, which grow total parameter count while keeping per-token inference compute roughly constant~\citep{lepikhin2021gshard,fedus2022switch}. This paradigm has driven a series of frontier MoE LLMs such as Mixtral~\citep{jiang2024mixtral}, DeepSeek-V3~\citep{deepseekai2024deepseekv3}, Qwen2.5~\citep{qwen2024qwen25}, and Kimi~K2~\citep{kimiteam2025kimik2}. These autoregressive models universally adopt \emph{token-choice} (TC) routing~\citep{lepikhin2021gshard,fedus2022switch}, where each token independently selects its preferred experts. TC is well suited to causal generation, since tokens are produced sequentially and a global view of all tokens is unavailable~\citep{muennighoff2024olmoe}. Recent DLMs have similarly adopted MoE for the same scaling benefits~\citep{zhu2025lladamoe,bie2025llada2}, but directly inherit TC routing from autoregressive systems without considering the unique structural properties of DLMs: non-causal attention and simultaneous generation of multiple tokens per denoising step. While \citet{ni2025openmoe2} also identify that TC routing may not be the best choice for DLMs, a systematic study of alternative routing paradigms and their unique advantages in diffusion language models remains lacking.

TC routing suffers from a well-known load imbalance problem: because tokens choose independently, some experts are overloaded while others remain idle. Mitigating this requires an auxiliary load-balancing loss~\citep{lepikhin2021gshard,fedus2022switch}, which consumes additional compute and introduces a competing gradient signal that can interfere with the primary language modeling objective~\citep{wang2024auxiliary}. Even so, load balance remains fragile with no hard guarantee~\citep{zhou2022mixture}. We argue that \emph{expert-choice} (EC) routing~\citep{zhou2022mixture}, where each expert selects a fixed number of tokens rather than the reverse, is the natural paradigm for DLMs. EC enforces deterministic capacity by design, eliminating load imbalance without any auxiliary loss. Crucially, DLMs process all tokens non-causally in each denoising step, unlike autoregressive models where future tokens are unavailable. This makes DLMs inherently compatible with EC, which requires a global view of all tokens to perform expert-side selection.

Beyond load balance, EC routing in DLMs unlocks a further opportunity: \emph{timestep-dependent computation}. DLM training and inference proceed through an explicit loop over denoising steps, each operating at a different masking ratio and thus facing a qualitatively different task. In TC routing, per-expert load is an emergent outcome of independent token choices and cannot be directly controlled. In EC routing, expert capacity is an external design variable, so we can vary it across timesteps to allocate more computation to the steps where it yields the highest return.

In this work, we make the following contributions:
\begin{itemize}
    \item We show that EC routing is strictly superior to TC routing for DLM MoE training, achieving better load balance, higher throughput, and faster convergence (\S\ref{sec:ec_vs_tc}).

    \item We introduce timestep-dependent expert capacity scheduling and evaluate multiple strategies under matched FLOPs. Per-timestep analysis reveals that tokens in low-mask-ratio contexts learn an order-of-magnitude faster, explaining why allocating more capacity to these steps yields the largest marginal return (\S\ref{sec:dynamic},~\S\ref{sec:mechanistic}).

    \item We show that existing pretrained TC DLMs can also benefit from EC: a simple router replacement enables faster convergence and improved accuracy during finetuning across diverse downstream tasks (\S\ref{sec:retrofit}).
\end{itemize}

%================================================================
\section{Background}
\label{sec:background}
%================================================================

\subsection{Diffusion Language Models}

Masked diffusion language models (DLMs)~\citep{austin2021structured,sahoo2024simple,nie2025llada} generate text by iteratively denoising an entire sequence, rather than producing tokens one at a time as in autoregressive models. A masking schedule $\gamma(t)\in[0,1]$ governs the fraction of positions replaced by \texttt{[MASK]} at noise level $t$. Decoding proceeds for $T$ steps:
\begin{equation}
  \mathbf{x}^{(T)} = [\texttt{MASK}]^L \;\xrightarrow{\;t=T\;}\; \mathbf{x}^{(T-1)} \;\rightarrow\; \cdots \;\xrightarrow{\;t=1\;}\; \mathbf{x}^{(0)},
\end{equation}
where at each step the model predicts all masked positions with bidirectional attention and unmasks a subset according to $\gamma(t)$. Without causal constraints, every forward pass processes the full $L$-token sequence. Notably, each step operates at a distinct masking ratio, presenting a qualitatively different denoising task; we revisit this property in Section~\ref{sec:dynamic}.

\subsection{Mixture-of-Experts Routing}

\begin{figure}[t]
  \centering
  \begin{subfigure}[t]{0.48\linewidth}
    \centering
    \includegraphics[width=\linewidth]{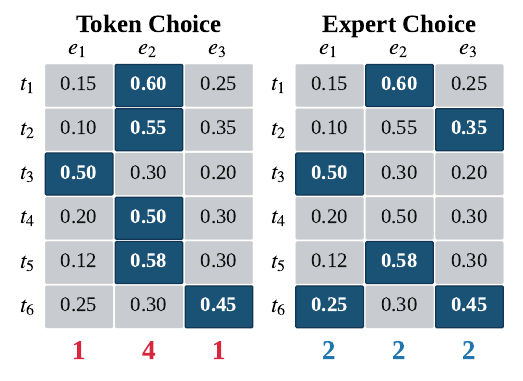}
    \caption{TC vs.\ EC routing.}
    \label{fig:routing_schematic}
  \end{subfigure}\hfill
  \begin{subfigure}[t]{0.48\linewidth}
    \centering
    \includegraphics[width=\linewidth]{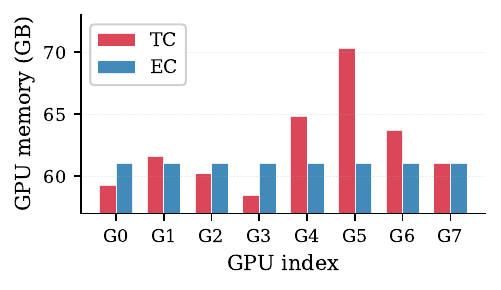}
    \caption{Per-GPU memory snapshot.}
    \label{fig:gpu_utilization_bg}
  \end{subfigure}
  \caption{\textbf{Left:} TC (top-1) vs.\ EC (capacity $c\!=\!2$) routing on a $6\!\times\!3$ gating score matrix. Both methods assign the same total of 6 token--expert pairs, but TC produces imbalanced per-expert loads (1/4/1) while EC guarantees uniform loads (2/2/2) by construction. \textbf{Right:} GPU memory snapshot during inference of LLaDA-2.0-mini (16B) with expert parallelism across 8 H100 GPUs. TC exhibits high variance (std 3.6\,GB) with one GPU using 70.3\,GB while others use ${\sim}$58--64\,GB. EC maintains perfectly uniform memory (std 0.0\,GB).}
  \label{fig:routing_overview}
\end{figure}

An MoE layer replaces a single FFN with $E$ parallel expert FFNs. Given $N$ tokens, a router computes a score matrix $\mathbf{S} \in \mathbb{R}^{N \times E}$, and a routing policy selects a sparse subset of token--expert pairs for computation. The two dominant policies differ in \emph{who selects whom} (Figure~\ref{fig:routing_schematic}).

\paragraph{Token-choice (TC)~\citep{fedus2022switch,lepikhin2021gshard}.} Each token selects its top-$k$ experts: $\mathcal{E}_i = \mathrm{TopK}_j(S_{i,j},\, k)$, yielding $\mathbf{y}_i = \sum_{j \in \mathcal{E}_i} g_{i,j}\, \mathrm{FFN}_j(\mathbf{x}_i)$ with normalized gates $g_{i,j}$. Per-expert load is uncontrolled: a capacity factor $\mathrm{CF}$ caps each expert at $\lceil \mathrm{CF} \cdot kN/E \rceil$ tokens and overflow is dropped. An auxiliary load-balancing loss encourages uniform utilization but remains fragile in practice.

\paragraph{Expert-choice (EC)~\citep{zhou2022mixture}.} The selection is inverted: each expert selects its top-$c$ tokens: $\mathcal{T}_j = \mathrm{TopC}_i(S_{i,j},\, c)$, yielding $\mathbf{y}_i = \sum_{j:\, i \in \mathcal{T}_j} g_{i,j}\, \mathrm{FFN}_j(\mathbf{x}_i)$. Every expert processes exactly $c$ tokens, so load balance is guaranteed by construction: no tokens are dropped and no auxiliary loss is needed. Setting $c = kN/E$ gives $Ec = kN$ total token--expert pairs, matching TC top-$k$ in total computation; varying $c$ scales compute proportionally.

%================================================================
\section{Expert-Choice as the Proper Routing Paradigm}
\label{sec:ec_vs_tc}
%================================================================

We first establish that expert-choice routing is consistently preferable to token-choice routing for DLM MoE models through controlled pretraining experiments. All models are trained from scratch with the same architecture, data, and hyperparameters; only the routing mechanism differs (full details in Appendix~\ref{app:setup}).

\subsection{Training Efficiency}

Figure~\ref{fig:loss_wallclock} compares training loss as a function of wall-clock time for EC and two representative TC configurations: dropless TC and capacity-bounded TC (cap=1.25). EC reaches loss 3.75 in 10.6h, approximately $2.0\times$ faster in wall-clock time than either TC variant (${\sim}$20h). Table~\ref{tab:throughput} in Appendix~\ref{app:throughput} confirms that this advantage stems from higher throughput: EC achieves 52.1 TFLOP/s/GPU, $1.5$--$2.1\times$ higher than all TC variants. Among capacity-bounded TC variants, throughput decreases with the capacity factor ($35.4 \to 27.0 \to 25.9$): a larger capacity raises the per-expert token ceiling, amplifying the straggler effect. TC~(dropless) is the slowest at 24.9, because unbounded per-expert load maximizes the straggler penalty. We ablate additional TC configurations (auxiliary loss variants, capacity factors) and provide per-step convergence analysis in Appendix~\ref{app:tc_ablations}. For capacity-bounded TC, the reported throughput uses the theoretical per-expert FLOPs; dropped tokens reduce the actual computation below this number. The gap between EC's measured throughput and TC's reported throughput is therefore a conservative lower bound on EC's true efficiency advantage.

\subsection{Load Balance and GPU Utilization}

Figure~\ref{fig:gpu_utilization_bg} visualizes the root cause. Under TC routing, per-GPU memory varies substantially (std 3.6\,GB), with one GPU using 70.3\,GB while others use ${\sim}$58--64\,GB. All GPUs must wait for the most loaded one. This imbalance is \emph{structural}: auxiliary load-balancing losses can mitigate but never eliminate it, because individual tokens still choose independently. EC removes this problem by construction, keeping all GPUs at uniform memory usage (std 0.0\,GB).

\noindent\textbf{Takeaway.} EC routing eliminates structural load imbalance, yielding higher throughput and faster wall-clock convergence than all TC variants (Appendix~\ref{app:tc_ablations}).

%================================================================
\section{Timestep-Adaptive Expert Capacity}
\label{sec:dynamic}
%================================================================

Beyond efficiency, EC routing unlocks a capability unavailable to TC: because expert capacity is an explicit hyperparameter rather than an emergent quantity, it can be \emph{scheduled} as a function of the denoising timestep. This raises a natural question: do all timesteps in a DLM truly need the same computation?

\subsection{Timestep-Dependent Capacity}

In DLMs, the masking ratio changes across denoising steps, presenting qualitatively different tasks at each timestep. Prior work on continuous diffusion has shown that different noise levels exhibit fundamentally different learning dynamics~\citep{kim2025speed,kim2025curriculum}, conflicting gradient contributions~\citep{hang2023efficient}, and schedule sensitivity~\citep{lin2024common}, but the relationship between masking ratio and computation demand in discrete masked diffusion remains unexplored. A natural question is whether all steps benefit equally from the same amount of computation. With EC routing, we can directly test this by varying the expert capacity as a function of the masking ratio.

However, it is unclear \emph{a priori} which masking ratios benefit most from additional computation, and three hypotheses are equally plausible. High mask ratios could be hardest, since most tokens are masked and extra experts might compensate for the information deficit. Low mask ratios could be hardest, since the few remaining masked tokens demand precise contextual reasoning that extra experts might refine. Intermediate ratios could be hardest, since around 50\% masking the task is neither trivially easy nor hopelessly underdetermined. Only experiments can distinguish them.

\subsection{Scheduling Strategies}

\begin{figure}[t]
  \centering
  \includegraphics[width=0.58\linewidth]{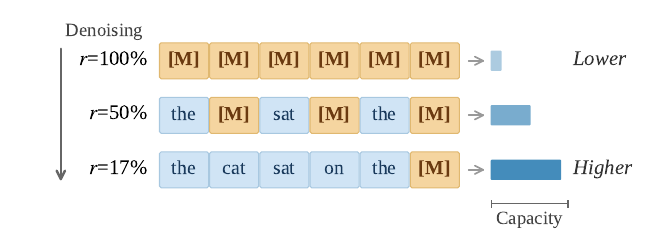}
  \caption{Linear-reverse scheduling: as mask ratio $r$ decreases during denoising, per-expert capacity increases, concentrating compute on the most consequential predictions.}
  \label{fig:dynamic_motivation}
\end{figure}

Let $r \in [0,1]$ denote the masking ratio at a given denoising step. We define a capacity function $k(r) = \mathrm{clamp}(k_{\min} + (k_{\max}-k_{\min}) \cdot s(r),\; k_{\min},\; k_{\max})$, where $s(r) \in [0,1]$ is a scheduler function. Figure~\ref{fig:dynamic_motivation} illustrates the idea for linear-reverse, and Table~\ref{tab:schedulers} lists all schedulers we consider. All are calibrated so that the average FLOPs across timesteps match the static EC baseline (constant $k = k_{\mathbb{E}}$), enabling fair comparison.
We treat $(k_{\min}, k_{\max})$ as a tunable pair, not as a tuned result. The FLOPs-matching constraint fixes $k_{\mathbb{E}}$ (Appendix~\ref{app:scheduler_flops}), leaving the spread $k_{\max}-k_{\min}$ to control how aggressively the schedule redistributes computation across timesteps. We use the widest spread our per-step memory budget allows, and leave a principled selection rule open.

\begin{table}[t]
\centering
\small
\caption{Scheduling strategies and final validation perplexity (30B tokens, OpenWebText). Expert capacity is $k(r) = k_{\min} + (k_{\max} - k_{\min}) \cdot s(r)$ with $k_{\min}{=}8$, $k_{\max}{=}32$; static baseline uses $k{=}20$. $\tilde{g}(r)$: normalized Gaussian centered at $r{=}0.5$ with $\sigma{=}0.22$ (Appendix~\ref{app:scheduler_flops}). All schedulers match in expected FLOPs.}
\label{tab:schedulers}
\begin{tabular}{@{}llcc@{}}
\toprule
Scheduler & $s(r)$ & Compute bias & PPL $\downarrow$ \\
\midrule
\textbf{Linear-Rev.} & $1-r$ & Low mask ratio & \textbf{36.5} \\
Static ($k$=20) & -- & -- & 37.1 \\
Cosine-Rev. & $\tfrac{1}{2}(1+\cos\pi r)$ & Low mask ratio & 37.2 \\
Gaussian & $\tilde{g}(r)$ & Intermediate & 37.3 \\
Linear & $r$ & High mask ratio & 37.5 \\
Gaussian-Rev. & $1-\tilde{g}(r)$ & Extremes & 37.6 \\
Cosine & $\tfrac{1}{2}(1-\cos\pi r)$ & High mask ratio & 37.6 \\
\bottomrule
\end{tabular}
\end{table}

\subsection{Scheduler Comparison on OpenWebText}
\label{sec:scheduler_comparison}

\paragraph{Setup.}
We train DLM MoE models on OpenWebText (${\sim}$9B tokens) for 30B tokens (multiple epochs). The static baseline uses constant $k{=}20$; dynamic variants use $k_{\min}{=}8$, $k_{\max}{=}32$, matching the static baseline in expected FLOPs ($\mathbb{E}[k]{=}20$). All other hyperparameters are shared (Appendix~\ref{app:setup}).

\begin{figure}[t]
  \centering
  \includegraphics[width=0.74\linewidth]{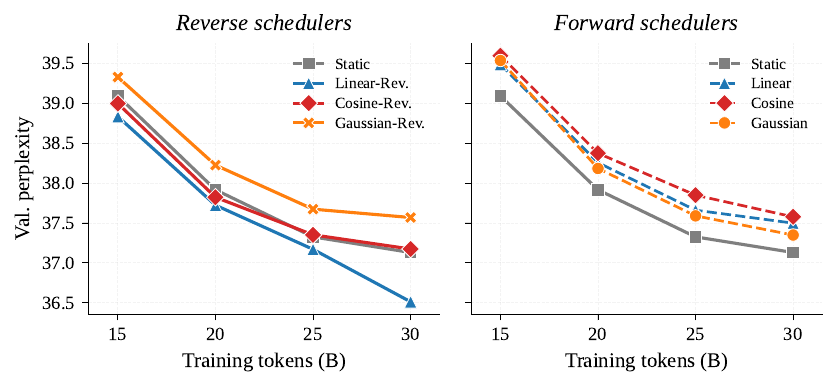}
  \caption{Scheduler comparison on OpenWebText (30B tokens, matched average FLOPs). \textbf{Left:}~reverse schedulers; \textbf{Right:}~forward schedulers. Reverse schedulers allocate more experts to low-mask-ratio steps and consistently outperform their forward counterparts.}
  \label{fig:scheduler_comparison}
\end{figure}

Figure~\ref{fig:scheduler_comparison} reports validation perplexity versus training tokens. \textbf{Linear-reverse consistently achieves the lowest perplexity} under matched FLOPs. More broadly, schedulers that allocate more computation to low-mask-ratio steps (linear-reverse, cosine-reverse) outperform those that favor high-mask-ratio steps (linear, cosine) or intermediate steps (Gaussian). This answers the question posed in \S\ref{sec:dynamic}:

\vspace{4pt}
\noindent\fcolorbox{black!25}{gray!6}{%
\begin{minipage}{0.95\linewidth}
\textbf{Takeaway:} Low-mask-ratio denoising steps benefit most from additional computation. Allocating more expert capacity to these steps consistently improves perplexity under matched FLOPs.
\end{minipage}}
\vspace{4pt}

\subsection{Scaling Validation: Pretraining at 8B-A1B Scale}

Our scheduler study is staged by scale. Section~\ref{sec:scheduler_comparison} runs the full bake-off across all six schedulers under matched FLOPs at small scale, where comparing every variant is tractable. This section then asks whether that choice transfers, since one 8B-A1B run costs about as much as the entire bake-off. The experiment therefore isolates the scheduling axis within EC and leaves the routing axis untouched: a TC baseline here would conflate the two, and \S\ref{sec:ec_vs_tc} settles EC versus TC on its own.

\paragraph{Setup.}
We pretrain two 8B-A1B (8B total, 1B active parameters) DLM MoE models on Nemotron-CC~\citep{su2024nemotroncc}. The static baseline uses constant $k{=}8$; the dynamic variant uses linear-reverse with $k_{\min}{=}2$, $k_{\max}{=}14$ ($\mathbb{E}[k]{=}8$, matched FLOPs). Architecture and all other hyperparameters are identical (Appendix~\ref{app:setup}).

\begin{figure}[t]
  \centering
  \includegraphics[width=0.80\linewidth]{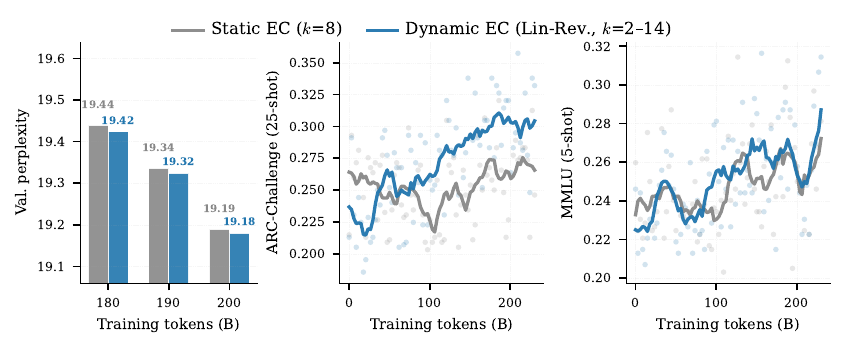}
  \caption{8B-A1B pretraining comparison: dynamic EC (linear-reverse, $k$=2--14) vs.\ static EC ($k$=8) on Nemotron-CC. Left: validation perplexity; Center: MMLU 5-shot accuracy; Right: ARC-Challenge 25-shot accuracy. Dynamic EC outperforms static EC at every checkpoint under matched average FLOPs. The linear-reverse schedule carries over from the small-scale bake-off in \S\ref{sec:scheduler_comparison} and was not re-evaluated at this scale; both curves are single runs.}
  \label{fig:downstream_1b}
\end{figure}

Figure~\ref{fig:downstream_1b} compares validation perplexity, MMLU~\citep{hendrycks2021mmlu} (5-shot), and ARC-Challenge~\citep{clark2018arc} (25-shot) accuracy as a function of training tokens. Dynamic linear-reverse EC consistently outperforms static EC across all three metrics, demonstrating that the scheduler advantage observed on OpenWebText transfers to larger scale and to downstream evaluations.
Both configurations are single training runs, since multi-seed pretraining at this scale is outside our compute budget. We state this as an explicit scope limitation and lean on consistency in place of significance testing: the gap holds at every checkpoint, not only at the end, and the scheduler ranking reproduces the six-way comparison of Table~\ref{tab:schedulers}.

\subsection{Mechanistic Analysis: Why Low-Mask-Ratio Steps Matter Most}
\label{sec:mechanistic}

Why does concentrating compute on low-mask-ratio steps help? We hypothesize that at low masking ratios, most tokens are already visible, providing rich context for the few remaining masked positions, so additional experts can meaningfully refine predictions. At high masking ratios, the model has little context and extra experts may yield diminishing returns. To test this hypothesis, we measure how fast the model learns at each masking ratio.

\paragraph{Setup.}
We partition the masking ratio into 4 equal bins ([0, 0.25), [0.25, 0.5), [0.5, 0.75), [0.75, 1.0)) and track the validation loss $\mathcal{L}_r$ within each bin $r$ over training. For each bin and training stage, we compute the \emph{convergence rate}
\begin{equation}
  \eta_r \;=\; -\frac{d\,\ln \mathcal{L}_r}{d\,t}\,,
  \label{eq:convergence_rate}
\end{equation}
via linear regression of $\ln(\mathcal{L}_r)$ against step $t$ (details in Appendix~\ref{app:convergence_rate}). Because different masking ratios produce losses at different scales, we operate in log-space so that $\eta_r$ measures the \emph{fractional} rate of loss decrease, enabling fair comparison across bins; larger values indicate faster learning.

\begin{figure}[t]
  \centering
  \includegraphics[width=0.74\linewidth]{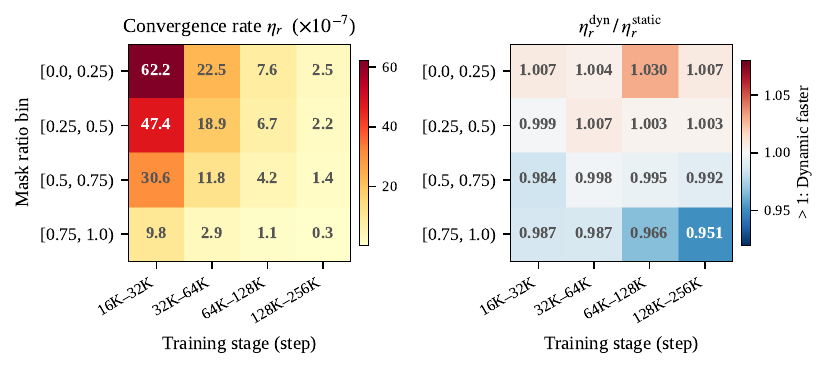}
  \caption{Mechanistic analysis of learning efficiency (8B-A1B, Nemotron-CC). \textbf{Left:}~Convergence rate $\eta_r$ (Eq.~\ref{eq:convergence_rate}) by mask-ratio bin and training stage for the static EC baseline; higher values indicate faster learning. Low-mask-ratio bins learn over $10\times$ faster. \textbf{Right:}~Ratio $\eta_r^{\text{dyn}} / \eta_r^{\text{static}}$; values ${>}1$ indicate dynamic EC learns faster. The advantage concentrates in low-mask-ratio bins.}
  \label{fig:mechanistic}
\end{figure}

\paragraph{Finding 1: Learning efficiency decreases monotonically with mask ratio.}
Figure~\ref{fig:mechanistic} (left) shows $\eta_r$ for the static EC baseline. Low-mask-ratio bins ([0, 0.25)) converge up to $7\times$ faster than high-mask-ratio bins ([0.75, 1.0)), and this gap widens over training. In other words, \textbf{low-mask-ratio steps are the model's primary learning frontier}.

\paragraph{Finding 2: Dynamic EC amplifies the advantage where it matters.}
Figure~\ref{fig:mechanistic} (right) shows $\eta_r^{\text{dyn}} / \eta_r^{\text{static}}$. Dynamic EC achieves higher convergence rates in low-mask-ratio bins (ratio ${>}1$) while slightly slower in high-mask-ratio bins (ratio ${<}1$). This trade-off is favorable: the gains concentrate in bins with the highest absolute $\eta_r$.

\paragraph{Summary.}
These findings suggest a consistent explanation: low-mask-ratio steps have the highest convergence rate ($\eta_r$ up to ${\sim}20\times$ that of high-mask-ratio steps), and linear-reverse allocates the most experts precisely to these steps, concentrating compute where the marginal return is highest. This also explains why Gaussian (midpoint-peaked) and linear (high-mask-favoring) schedulers underperform: they invest compute in regions with diminishing or near-zero marginal returns.

This explanation has limits worth stating. The convergence rate accounts for the \emph{direction} of the effect, that schedulers favoring low mask ratios beat those favoring high or intermediate ones. It does not account for the residual gap between the two reverse schedulers. Cosine-reverse also favors low-mask-ratio steps but is \emph{more aggressive} than linear-reverse, allocating more capacity near $r{=}0$ and less near $r{=}1$, yet the two show nearly identical per-bin convergence rates (Appendix~\ref{app:lin_vs_cos}) while differing in perplexity. Our working hypothesis is that cosine-reverse starves high-mask-ratio steps below a useful threshold, though we have no direct test and leave the question open. One further caveat: $\eta_r$ is a \emph{fractional} rate and measures optimization speed, not task importance. Appendix~\ref{app:convergence_rate} discusses this distinction, and \S\ref{sec:dynamic} establishes the perplexity conclusion on its own.

%================================================================
\section{Retrofitting Pretrained TC DLMs}
\label{sec:retrofit}
%================================================================

The preceding sections demonstrate the advantages of EC and dynamic EC when training from scratch. A practical question remains: can existing pretrained TC DLM models also benefit? Our retrofit is a low-cost router swap followed by standard finetuning, which recovers and exceeds TC accuracy while retaining a persistent decode speedup. It proceeds in two distinct stages.

\paragraph{Stage 1: conversion.} We replace the token-choice gate with an expert-choice gate. The change is purely in the selection rule applied to the same score matrix $\mathbf{S}$: TC takes a per-row top-$k$, EC takes a per-column top-$c$. The router projection keeps its pretrained TC values and is not randomly reinitialized, and expert FFNs, embeddings, attention, and normalization layers are all unchanged. No architectural changes are required beyond the router itself.

\paragraph{Stage 2: finetuning.} Each converted model then undergoes standard task-specific finetuning (\S\ref{sec:sft}), which adapts the experts to the new selection rule. Dynamic EC introduces no additional parameters; only $c$ becomes a function of the masking ratio (\S\ref{sec:dynamic}), so its trainable parameter set is identical to static EC. The results below reflect both stages, not the swap alone.

\subsection{Task-Specific Finetuning}
\label{sec:sft}

We finetune LLaDA-MoE, a pretrained TC DLM, on three tasks spanning different capability dimensions: \textbf{GSM8K}~\citep{cobbe2021gsm8k} (mathematical reasoning, trained on GSM8K-AUG-NL~\citep{deng2024gsm8kaug}), \textbf{HumanEval/HumanEval+}~\citep{chen2021humaneval,liu2023evalplus} (code generation, trained on OpenCodeInstruct~\citep{ahmad2025opencodeinstruct}), and \textbf{MedQA}~\citep{jin2020medqa} (medical knowledge question answering).
For each task, we compare three configurations: the original TC model, the converted EC model (static), and the converted dynamic EC model (linear-reverse scheduler). All configurations use identical finetuning hyperparameters.

\begin{figure}[t]
  \centering
  \includegraphics[width=0.78\linewidth]{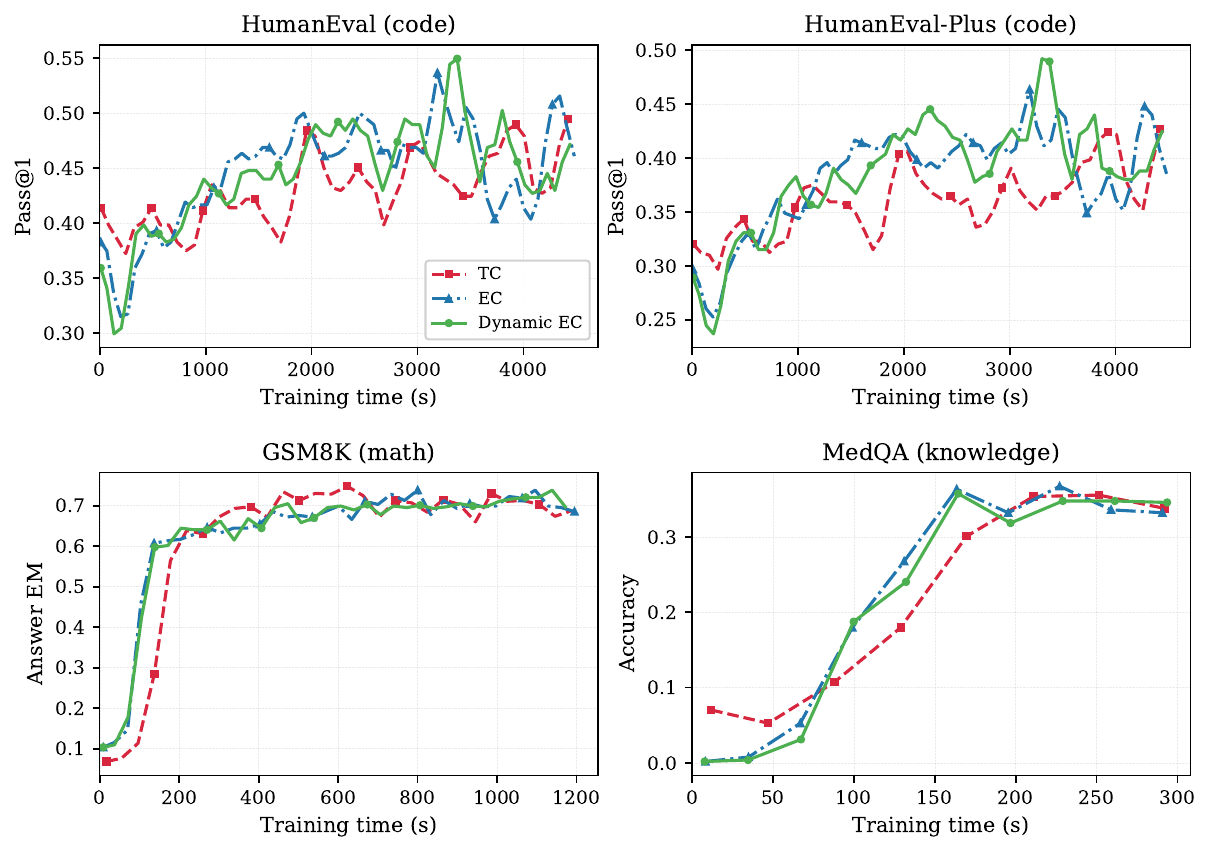}
  \caption{Retrofitting a pretrained TC DLM (LLaDA-MoE) to EC and dynamic EC across four benchmarks. The $t{=}0$ point of each EC curve is the post-swap, pre-finetuning baseline: the router swap alone costs a few points (HumanEval $0.42 \to 0.38$, HumanEval+ $0.33 \to 0.30$ Pass@1), which finetuning then recovers and exceeds. EC converges notably faster than TC in wall clock time while reaching comparable final accuracy, and dynamic EC achieves slightly higher final accuracy than both TC and static EC. See Appendix~\ref{app:sft_samples} for results plotted against training samples.}
  \label{fig:retrofit_tasks}
\end{figure}
Figure~\ref{fig:retrofit_tasks} reports accuracy versus wall-clock time. Across all three tasks, EC converges notably faster than TC while reaching comparable final accuracy, confirming that the router conversion does not degrade model quality. Dynamic EC (linear-reverse) further improves, achieving the highest average accuracy across the four benchmarks (54.9\% vs.\ 53.6\% for static EC and 52.6\% for TC).

Finetuning is a required part of the retrofit, not an optional refinement. The $t{=}0$ point of each EC curve in Figure~\ref{fig:retrofit_tasks} is the zero-shot accuracy immediately after conversion, before any finetuning step, and the swap alone causes a small initial drop (HumanEval $0.42 \to 0.38$, HumanEval+ $0.33 \to 0.30$ Pass@1). This is expected, since the expert FFNs were optimized under TC's load distribution and EC re-routes them with a different per-step token mixture. Finetuning exists precisely to adapt the experts to the new rule, and updates all parameters, including the EC router and every expert FFN, under the same learning rate and schedule as the TC baseline. After finetuning, EC improves over TC on three of the four benchmarks (HumanEval $+1.6$, HumanEval+ $+2.3$, MedQA $+1.1$, GSM8K $-1.0$); we read the GSM8K gap as typical finetuning seed variance, not a systematic property of EC routing.

Table~\ref{tab:sft_summary} quantifies both the peak accuracy and the evaluation decode time at the best checkpoint. Across all four benchmarks, EC and dynamic EC match or exceed TC in peak accuracy, while requiring significantly less time per evaluation round. This speedup stems from the same deterministic load balancing that benefits training: at inference time, EC eliminates the straggler effects of TC routing, yielding $1.3$--$1.5\times$ faster decoding.

\begin{table}[t]
\centering
\small
\caption{Peak accuracy (\%) and evaluation decode time (s) for TC-to-EC retrofitting. Peak values are the best scores during SFT (within the time window of Figure~\ref{fig:retrofit_tasks}). EC and dynamic EC achieve comparable or higher accuracy while decoding $1.3$--$1.5\times$ faster.}
\label{tab:sft_summary}
\setlength{\tabcolsep}{4pt}
\begin{tabular}{@{}lcccccccccc@{}}
\toprule
& \multicolumn{2}{c}{HumanEval} & \multicolumn{2}{c}{HumanEval+} & \multicolumn{2}{c}{GSM8K} & \multicolumn{2}{c}{MedQA} & \multicolumn{2}{c}{Avg} \\
\cmidrule(lr){2-3} \cmidrule(lr){4-5} \cmidrule(lr){6-7} \cmidrule(lr){8-9} \cmidrule(lr){10-11}
Routing & Pass@1 & Time & Pass@1 & Time & Acc & Time & Acc & Time & Acc & Time \\
\midrule
TC         & 53.9 & 1369 & 46.1 & 1369 & \textbf{74.8} & 672  & 35.6 & 1885 & 52.6 & 1324 \\
EC         & 55.5 & \textbf{1008} & 48.4 & \textbf{1008} & 73.8 & \textbf{459} & \textbf{36.7} & \textbf{1371} & 53.6 & \textbf{962} \\
Dynamic EC & \textbf{58.6} & 1056 & \textbf{51.6} & 1056 & 73.8 & 461  & 35.7 & 1378 & \textbf{54.9} & 988 \\
\bottomrule
\end{tabular}
\end{table}

These results demonstrate that existing deployed TC DLMs can benefit from EC routing with minimal modification: replace the router, finetune briefly, and obtain both faster convergence, faster inference, and (with dynamic scheduling) improved final quality.

\subsection{Inference-Time Properties}
\label{sec:inference_properties}

EC selects tokens from whatever pool is visible at each forward pass, which raises two deployment questions.

\paragraph{Single-sample decoding.} With batch size 1 and sequence length $L$, each expert still selects its top-$c$ tokens from the same $L$-token pool, so per-expert load remains uniform by construction within the sample. Per-step compute is bounded by $k_{\max}$, a constant fixed by the schedule (\S\ref{sec:dynamic}), so timestep-dependent capacity does not introduce unbounded per-step latency.

\paragraph{Block-wise and semi-autoregressive decoding.} Within a block, denoising is locally non-causal and EC applies directly. Across blocks, causal masking precludes a global expert-side view, so full-sequence EC does not apply. The natural composition is EC within a block and TC across blocks, which we leave to future work.

\vspace{4pt}
\noindent\textbf{Takeaway.} Pretrained TC DLMs can be retrofitted to EC by replacing only the router, achieving faster convergence, faster decoding, and improved accuracy.

%================================================================
\section{Related Work}
%================================================================

\paragraph{Diffusion language models.}
Discrete diffusion models for text generation have progressed from early foundations~\citep{austin2021structured,lou2024discrete,sahoo2024simple} to large-scale systems that rival autoregressive LLMs~\citep{nie2025llada,bie2025llada2,ye2025dream,arriola2025block}. Recent work has explored adapting pretrained AR models into diffusion LMs~\citep{gong2025scaling,liu2025wedlm,liu2025tidar}, and commercial deployments such as Mercury~\citep{khanna2025mercury} demonstrate the practical viability of diffusion-based generation. Sparse MoE variants of DLMs have also emerged~\citep{zhu2025lladamoe,ni2025openmoe2}. Our work is complementary: we do not propose a new diffusion process but rather study how MoE routing interacts with the denoising structure.

\paragraph{Mixture-of-experts.}
Sparse MoE architectures scale model capacity with sublinear compute cost~\citep{shazeer2017outrageously,lepikhin2021gshard,fedus2022switch,zoph2022stmoe}. A central challenge is load balancing: token-choice routing produces uneven expert loads, and mitigations range from auxiliary losses~\citep{shazeer2017outrageously,fedus2022switch}, optimal assignment~\citep{lewis2021base}, auxiliary-loss-free bias correction~\citep{wang2024auxiliary,deepseekai2024deepseekv3}, to ReLU-based differentiable routing~\citep{wang2025remoe} and threshold-based routing~\citep{sun2026expert}. Expert-choice routing~\citep{zhou2022mixture} sidesteps the problem entirely by letting each expert select a fixed number of tokens, guaranteeing perfect balance by construction, but was previously studied only on encoder-decoder models. Recent MoE LLMs adopt fine-grained experts and shared-expert isolation~\citep{dai2024deepseekmoe,jiang2024mixtral,deepseekai2024deepseekv3}, trained at scale with frameworks such as Megatron-MoE~\citep{yan2026megatronmoe}. We are the first to systematically study EC routing and timestep-dependent capacity scheduling in diffusion language models.

\paragraph{Timestep-dependent computation.}
Our finding that low-mask-ratio steps benefit most from additional computation aligns with, rather than contradicts, prior analyses of continuous diffusion, under the correspondence between low noise and low mask ratio. \citet{kim2025curriculum} report that denoising tasks at smaller $t$ are harder to learn and build a curriculum advancing from high to low noise, while \citet{kim2025speed} find that the high-noise regime converges easily and is oversampled whereas the lower-noise band is undersampled, and recommend reallocating training toward it. Our linear-reverse capacity schedule points in the same direction. In the MoE setting, EC-DIT~\citep{sun2025ecdit} and DiffMoE~\citep{shi2025diffmoe} are concurrent work on vision diffusion transformers: the former shows that EC routing implicitly adapts to timesteps, the latter learns a per-timestep capacity predictor. Ours is the first systematic study of EC routing and timestep-dependent capacity in discrete masked diffusion \emph{language} models, with the per-mask-ratio mechanistic analysis (\S\ref{sec:mechanistic}) as its principal novel component. We use hand-designed schedules instead of a learned capacity predictor, since a fair learned comparison needs substantial additional architecture and stability work; ours serve as a mechanistic baseline that makes the direction of the effect explicit. Learning the schedule is left to future work.

We discuss adaptive computation and timestep-adaptive methods in Appendix~\ref{app:additional_related_work}.

%================================================================
\section{Conclusion}
%================================================================

We have shown that expert-choice routing is consistently preferable to token-choice routing for DLM MoE models, providing higher throughput through deterministic load balancing and enabling timestep-dependent expert capacity scheduling. Our mechanistic analysis reveals that low-mask-ratio denoising steps have an order of magnitude higher learning efficiency, explaining why the linear-reverse scheduler outperforms all alternatives under matched FLOPs. We further demonstrate that pretrained TC models can be retrofitted to EC by simply replacing the router. Together, these results suggest that in diffusion language models, \textbf{computation should be treated as an adaptive policy rather than a fixed architectural constant}. Appendix~\ref{app:future} outlines directions that follow from this view.

% \section*{Author Contributions}

% \section*{Acknowledgments}
% Use unnumbered first level headings for the acknowledgments. All

\newpage
\section*{Ethics Statement}
This work studies routing and computation scheduling in MoE diffusion language models. Our experiments use publicly available datasets (Nemotron-CC, OpenWebText, GSM8K, HumanEval, MedQA) and do not involve human subjects or private data. The methods we propose are general architectural improvements that do not introduce new risks beyond those inherent to large language models. We encourage responsible deployment practices when applying these techniques to production systems.

\bibliography{colm2026_conference}
\bibliographystyle{colm2026_conference}

\newpage
\appendix
\section*{Appendix}

\section*{Contents of the Appendix}
The appendix includes the following contents:
\begin{itemize}
    \item Sec.~\ref{app:limitations} discusses limitations.
    \item Sec.~\ref{app:future} outlines future directions.
    \item Sec.~\ref{app:llm_disclosure} discloses the use of large language models.
    \item Sec.~\ref{app:token_coverage} analyzes token coverage in expert-choice routing.
    \item Sec.~\ref{app:setup} provides full experimental setup details for all experiments.
    \item Sec.~\ref{app:tc_ablations} ablates TC routing variants (auxiliary loss, capacity factor, throughput).
    \item Sec.~\ref{app:convergence_rate} details the convergence rate computation, compares linear-reverse vs.\ cosine-reverse, and reports convergence-rate trajectories across training.
    \item Sec.~\ref{app:scheduler_flops} defines all schedulers and proves FLOPs equivalence.
    \item Sec.~\ref{app:sft_samples} presents additional SFT retrofitting results.
    \item Sec.~\ref{app:additional_related_work} discusses adaptive computation and timestep-adaptive methods.
\end{itemize}

\section{Limitations}
\label{app:limitations}

The capacity schedules explored in this work are hand-designed functions (linear, cosine, Gaussian and their reverses). While our mechanistic analysis explains why linear-reverse is effective, the optimal schedule may depend on model scale, dataset, and task. A natural extension is to replace hand-designed schedules with \emph{learned} ones: for example, a lightweight capacity predictor trained end-to-end (as in DiffMoE~\citep{shi2025diffmoe} for vision), or a policy learned via reinforcement learning that adapts capacity based on training signals. We leave these directions to future work.

\section{Future Directions}
\label{app:future}

\paragraph{Learned capacity.} The schedules studied here are hand-designed. Replacing them with a learned capacity predictor or a policy trained by reinforcement learning is a natural extension, and the per-bin analysis of \S\ref{sec:mechanistic} predicts the direction such a policy should converge to.

\paragraph{Continued pretraining before finetuning.} Our retrofit deliberately measures a lower bound on EC's benefit by swapping the router and finetuning directly. Inserting a short continued-pretraining stage on general text before task finetuning may recover more of the original model's quality, at the cost of a less direct attribution to the router swap.

\paragraph{Inference-time mechanisms unique to EC in DLMs.} EC's controllable capacity, combined with non-causal denoising, enables mechanisms available in neither autoregressive MoE nor TC-routed DLMs: timestep-aware speculative decoding, where low-mask steps verify drafts produced at higher mask ratios and $c$ tunes verification strictness; per-prompt adaptive capacity, where $c$ is set at inference from predictive uncertainty; and composition with adaptive parallel decoding~\citep{israel2025adaptive}, where deterministic load balance simplifies variable-batch bookkeeping.

\paragraph{Block-wise diffusion.} Extending the EC-within-block, TC-across-block composition of \S\ref{sec:inference_properties} to block diffusion models~\citep{arriola2025block} is a concrete next step toward semi-autoregressive deployment.

\section{Use of Large Language Models}
\label{app:llm_disclosure}

All research ideas, experimental design, theoretical analysis, and scientific conclusions were conceived and developed by the authors. LLM assistance was used in the following limited capacities: (1) writing and debugging experiment and plotting code, and (2) drafting and polishing prose in the manuscript. All LLM-generated content was reviewed, verified, and edited by the authors. No LLM was used to originate research ideas or generate experimental data.

\section{Token Coverage in Expert-Choice Routing}
\label{app:token_coverage}

A natural concern with expert-choice routing is that some tokens may not be selected by any routed expert, potentially causing information loss. We address this with both empirical measurements and a probabilistic argument.

\begin{figure}[h]
  \centering
  \includegraphics[width=0.85\linewidth]{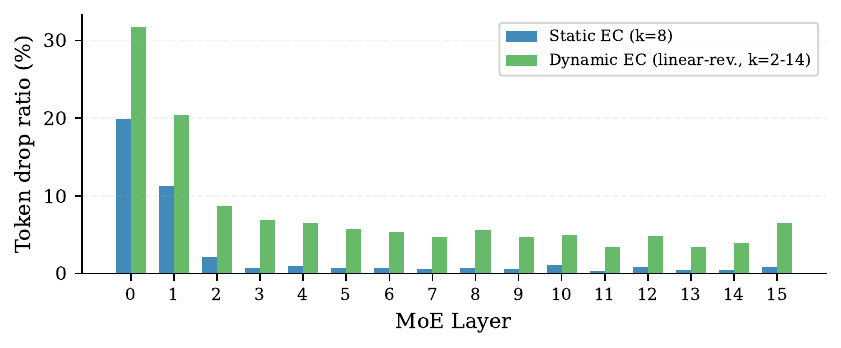}
  \caption{Per-layer token drop ratio at step 300K for static and dynamic EC (8B-A1B, Nemotron-CC). Middle layers drop $<$1.1\% of tokens under static EC. Dynamic EC has higher drop ratios due to reduced capacity at high-mask-ratio steps.}
  \label{fig:token_drop}
\end{figure}

\paragraph{Per-layer token drop ratio.}
Figure~\ref{fig:token_drop} reports the fraction of tokens not selected by any routed expert at each MoE layer, measured at step 300K (5-point average). For static EC ($k{=}8$), middle layers (2--14) drop fewer than 1.1\% of tokens, with a mean of 2.7\% across all layers. Dynamic EC (linear-reverse, $k{=}2$--$14$) has a higher mean drop ratio (8.0\%) because low-capacity steps ($k{=}2$) naturally leave more tokens unrouted. Both variants show elevated drop ratios at layer 0 (20--32\%), likely because the first router has not yet developed strong token-expert affinities.

\paragraph{Why information is not lost.}
The primary reason is architectural. The model includes shared experts that process \emph{all} tokens unconditionally at every layer, regardless of routing decisions. Even when a token is not selected by any routed expert in a given layer, it is still processed by the shared expert FFN, so a drop removes a token from the \emph{routed} pathway, not from computation altogether.

We also bound how often a token can be dropped across the full depth. Router decisions at different layers are correlated through the residual stream, so we do not assume that per-layer drop events are independent. Under the worst case of perfectly correlated drops, the probability that a token is unrouted at every layer is upper-bounded by the per-layer marginal drop rate, that is, ${\sim}2.7\%$ for static EC and ${\sim}8.0\%$ for dynamic EC. This bound is far looser than what an independence assumption would give, and we prefer it because the assumption is unverified. Combined with the shared-expert argument above, even this worst case does not imply information loss. Measuring the empirical cross-layer correlation of drop events directly is left to future work.

\section{Experimental Setup}
\label{app:setup}

All experiments use the Megatron-LM framework with SwiGLU expert activations and AdamW optimizer ($\beta_1{=}0.9$, $\beta_2{=}0.95$, WSD learning rate schedule).

\paragraph{EC vs.\ TC comparison (\S\ref{sec:ec_vs_tc}).}
16 Transformer layers, hidden size 2048, 64 fine-grained experts (ffn hidden size 1280), 2 shared experts. Training data: Nemotron-CC~\citep{su2024nemotroncc} (50/50 blend of High-Quality and Diverse-QA splits). Learning rate $2{\times}10^{-4}$. TC and EC models are identical except for the routing mechanism; all other hyperparameters are shared.

\paragraph{Scheduler comparison on OpenWebText (\S\ref{sec:dynamic}).}
16 Transformer layers, hidden size 512, 16 attention heads, 512 fine-grained experts (ffn hidden size 384), 2 shared experts (shared ffn hidden size 768). Sequence length 513, global batch size 256, trained for 30B tokens on OpenWebText (${\sim}$9B tokens, multiple epochs). Learning rate $2{\times}10^{-4}$. Dynamic variants: $k_{\min}{=}8$, $k_{\max}{=}32$; static baseline: constant $k{=}20$.

\paragraph{8B-A1B pretraining (\S\ref{sec:dynamic}).}
16 Transformer layers, hidden size 2048, 16 attention heads, 64 fine-grained experts (ffn hidden size 1280), 2 shared experts. Sequence length 2049, global batch size 288. Training data: Nemotron-CC (50/50 High-Quality / Diverse-QA). Learning rate $2{\times}10^{-4}$. Static baseline: constant $k{=}8$; dynamic variant: linear-reverse with $k_{\min}{=}2$, $k_{\max}{=}14$.

\section{TC Routing Ablations}
\label{app:tc_ablations}

\begin{figure}[h]
  \centering
  \begin{subfigure}[t]{0.48\linewidth}
    \centering
    \includegraphics[width=\linewidth]{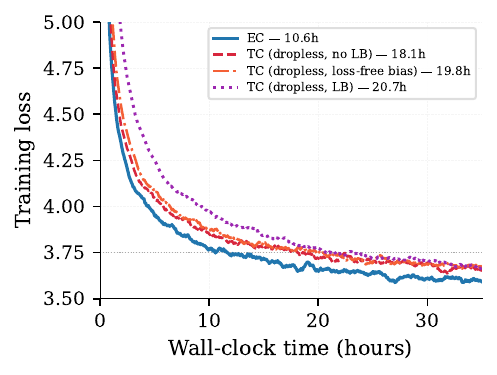}
    \caption{Auxiliary loss ablation.}
    \label{fig:dropless_tc}
  \end{subfigure}\hfill
  \begin{subfigure}[t]{0.48\linewidth}
    \centering
    \includegraphics[width=\linewidth]{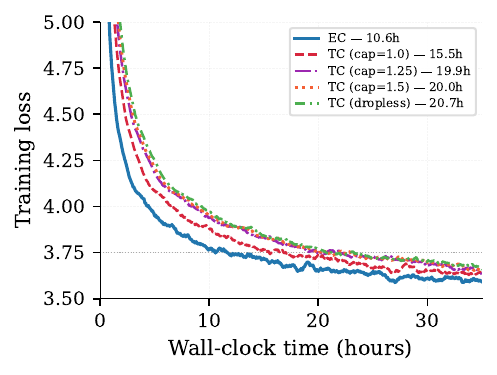}
    \caption{Capacity factor ablation.}
    \label{fig:capacity_ablation}
  \end{subfigure}
  \caption{TC routing ablations. \textbf{(a)}~Removing or replacing the auxiliary load-balancing loss does not close the gap with EC ($1.7\times$ faster). \textbf{(b)}~Increasing the capacity factor slows TC further due to padding overhead ($2.0\times$ faster for EC vs.\ TC cap=1.5).}
  \label{fig:tc_ablations}
\end{figure}

We ablate two aspects of TC routing to identify the source of EC's convergence advantage. Figure~\ref{fig:tc_ablations} summarizes both experiments. In all plots, wall-clock time is measured as cumulative per-iteration training time, excluding evaluation and checkpointing.

\subsection{Auxiliary Loss}
\label{app:aux_loss_ablation}

The TC baselines in \S\ref{sec:ec_vs_tc} use an auxiliary load-balancing loss to encourage uniform expert utilization. A natural question is whether this auxiliary loss, rather than the routing mechanism itself, is the primary source of TC's slower convergence, since the auxiliary loss introduces additional gradient signals that may interfere with the main language modeling objective.

To rule out this confound, we compare EC against two \emph{dropless} TC variants that eliminate or replace the auxiliary loss:

\begin{itemize}
  \item \textbf{TC (dropless, no LB):} Dropless TC routing with no auxiliary load-balancing loss at all. Experts process all assigned tokens regardless of load skew.
  \item \textbf{TC (dropless, loss-free bias):} Dropless TC routing with the auxiliary-loss-free balancing strategy of \citet{wang2024auxiliary}, which adjusts expert selection via learned bias terms without introducing any auxiliary gradient signal. We use the recommended bias update rate of $0.001$.
\end{itemize}

As shown in Figure~\ref{fig:dropless_tc}, EC still converges $1.7\times$ faster than the slowest dropless TC variant. Removing the auxiliary loss entirely or replacing it with a loss-free alternative does not close the gap with EC. This confirms that the auxiliary load-balancing loss is not the primary factor behind TC's slower convergence; rather, the dominant bottleneck is the fundamental load imbalance inherent to token-choice routing.

\begin{figure}[h]
  \centering
  \includegraphics[width=0.48\linewidth]{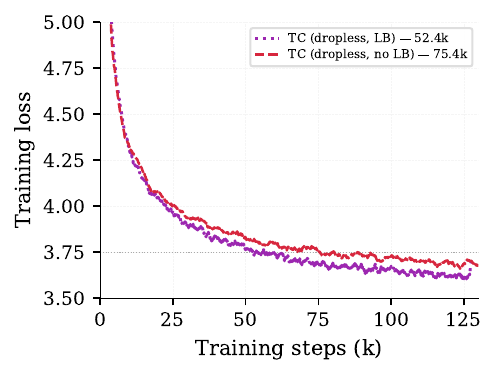}
  \caption{Per-step training loss. The auxiliary load-balancing loss improves per-step convergence (TC dropless LB reaches loss 3.75 in fewer steps), but this gain is offset by reduced throughput (Table~\ref{tab:throughput}), resulting in slower wall-clock convergence (Figure~\ref{fig:dropless_tc}).}
  \label{fig:perstep_lb}
\end{figure}

Interestingly, comparing TC (dropless, LB) and TC (dropless, no LB) reveals a tension between per-step convergence and throughput (Figure~\ref{fig:perstep_lb}). The auxiliary loss does improve per-step optimization: TC (dropless, LB) reaches loss 3.75 in only 50.6k steps versus 78.6k steps for TC (dropless, no LB), a 36\% reduction in required iterations. However, the auxiliary loss also exacerbates load imbalance at the system level, reducing throughput from 44.4 to 24.9 TFLOP/s/GPU (Table~\ref{tab:throughput}), a $1.78\times$ slowdown per step. The throughput penalty outweighs the per-step gain, so TC (dropless, no LB) reaches loss 3.75 at 18.1h versus 20.7h for TC (dropless, LB). EC sidesteps this tradeoff entirely: it achieves both perfect load balance (maximizing throughput) and efficient per-step optimization, reaching the same loss in just 10.6h.

\subsection{Capacity Factor}
\label{app:capacity_ablation}

The capacity factor controls the maximum number of tokens each expert can process per step. A larger capacity factor reduces token dropping, improving per-step data utilization. However, it also raises the maximum allowed load per expert to capacity factor $\times$ $N/E$, amplifying the straggler effect: the most overloaded expert takes longer to finish, and all GPUs must wait. Figure~\ref{fig:capacity_ablation} compares EC against TC with capacity factors 1.0, 1.25, and 1.5. Increasing the capacity factor slows wall-clock convergence: TC~(cap=1.0) reaches loss 3.75 at 15h, TC~(cap=1.25) at 19h, and TC~(cap=1.5) at 20h. EC converges at 10h in all comparisons, achieving up to $2.0\times$ speedup. The marginal benefit of retaining more tokens per step does not compensate for the increased per-step overhead, widening the gap with EC.

\subsection{Throughput Analysis}
\label{app:throughput}

We report training throughput in TFLOP/s/GPU based on the forward-pass FLOPs per step:
\begin{equation}
  \text{Throughput} = \frac{F_{\text{fwd}}}{t_{\text{step}} \times 10^{12} \times N_{\text{GPU}}},
\end{equation}
where $t_{\text{step}}$ is the measured wall-clock time per training step and $F_{\text{fwd}}$ is the theoretical forward-pass floating-point operations, computed as $F_{\text{fwd}} = 2 \cdot B \cdot L_{\text{seq}} \cdot N_{\text{layers}} \cdot d^2 \cdot (\text{attention} + \text{MLP} + \text{logit terms})$, with the MLP term accounting for all routed experts ($k \cdot d_{\text{ffn}} / d$) and shared experts separately~\citep{narayanan2021efficient}. Since all models share the same architecture, $F_{\text{fwd}}$ is identical across runs; differences in throughput reflect only hardware utilization efficiency.

\begin{table}[h]
\centering
\small
\caption{Average training throughput (TFLOP/s/GPU) over the first 35h. Higher is better. EC achieves $1.5$--$2.1\times$ higher throughput than all TC variants.}
\label{tab:throughput}
\begin{tabular}{@{}lcc@{}}
\toprule
Routing & Throughput & Relative to EC \\
\midrule
\textbf{EC}                          & \textbf{52.1} & 1.00$\times$ \\
\midrule
TC (dropless, no LB)                 & 44.4 & 0.85$\times$ \\
TC (dropless, loss-free bias)        & 38.6 & 0.74$\times$ \\
TC (cap=1.0, LB)                     & 35.4 & 0.68$\times$ \\
TC (cap=1.25, LB)                    & 27.0 & 0.52$\times$ \\
TC (cap=1.5, LB)                     & 25.9 & 0.50$\times$ \\
TC (dropless, LB)                    & 24.9 & 0.48$\times$ \\
\bottomrule
\end{tabular}
\end{table}
Table~\ref{tab:throughput} reports the average throughput over the first 35 hours of training. Two trends emerge:

\paragraph{EC uniformly outperforms all TC variants.} EC achieves 52.1 TFLOP/s/GPU, $1.2\times$--$2.1\times$ higher than every TC configuration. This advantage stems directly from deterministic load balancing: all GPUs process exactly the same number of tokens per step, eliminating idle time.

\paragraph{Load imbalance is the dominant bottleneck, not auxiliary loss.} Among the dropless TC variants, removing the auxiliary loss (no LB, 44.4) or replacing it with loss-free bias (38.6) does not close the gap with EC. The auxiliary loss itself incurs negligible computational cost; the throughput differences reflect varying degrees of load imbalance across configurations. For capacity-bounded TC, throughput decreases monotonically with capacity factor ($35.4 \to 27.0 \to 25.9$): a larger capacity factor raises the per-expert token ceiling, amplifying the straggler effect where all GPUs wait for the most loaded one. Note that $F_{\text{fwd}}$ assumes every token is routed to $k$ experts; capacity-bounded TC drops overflow tokens, so its \emph{actual} computation is lower than $F_{\text{fwd}}$, meaning the reported throughput for these variants is an \emph{upper bound}.

In summary, EC's throughput advantage is structural: it eliminates load variance at the routing level rather than mitigating it with auxiliary objectives or capacity bounds.

\section{Convergence Rate Computation}
\label{app:convergence_rate}

\paragraph{Computation.}
During validation, we record the per-token cross-entropy loss $\mathcal{L}_r(t)$ separately for each masking-ratio bin $r$, weighted by token count. We estimate $\eta_r$ by fitting a linear model to $\ln \mathcal{L}_r$ within each training stage via least-squares regression:
\begin{equation}
  \hat{\eta}_r = -\underset{b_1}{\operatorname{argmin}} \sum_{t \in \text{stage}} \big(\ln \mathcal{L}_r(t) - b_1 t - b_0\big)^2.
\end{equation} Training is divided into geometrically spaced stages (16K--32K, 32K--64K, 64K--128K, 128K--256K steps) so that later stages, where loss changes more slowly, use proportionally longer windows.

\paragraph{Interpretation.}
By the chain rule, $\eta_r = -(1/\mathcal{L}_r) \cdot d\mathcal{L}_r/dt$ measures the \emph{fractional} rate of loss decrease per step, independent of the absolute loss magnitude. This allows comparison across bins with different loss scales. Since different bins have different irreducible loss floors, $\eta_r$ describes optimization speed rather than task importance. In particular, a bin with a smaller absolute loss tends to show a larger fractional rate for the same absolute improvement, so ``learns faster'' in this sense should not be read as ``is more important''. The scheduler comparison in \S\ref{sec:dynamic} is what independently establishes that allocating capacity to low-mask-ratio steps improves end-task perplexity.

\subsection{Linear-Reverse vs.\ Cosine-Reverse}
\label{app:lin_vs_cos}

\begin{figure}[h]
  \centering
  \includegraphics[width=\linewidth]{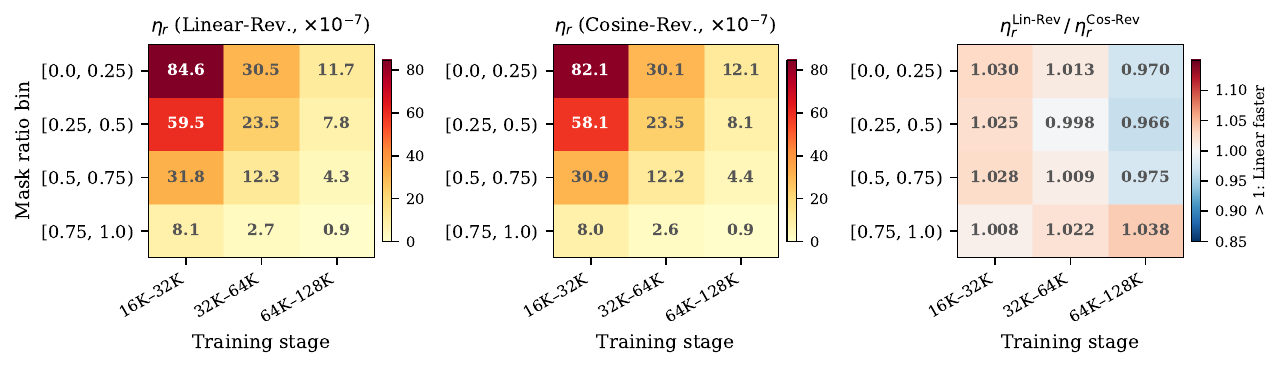}
  \caption{Convergence rate comparison between linear-reverse and cosine-reverse (both from H100 runs, evaluated up to 128K steps). \textbf{Left/Center:} per-bin $\eta_r$ for each scheduler. \textbf{Right:} ratio $\eta_r^{\text{Lin-Rev}} / \eta_r^{\text{Cos-Rev}}$; values near 1.0 indicate no meaningful difference. The two schedulers have nearly identical per-bin learning dynamics.}
  \label{fig:lin_vs_cos}
\end{figure}

Figure~\ref{fig:lin_vs_cos} compares the per-bin convergence rate $\eta_r$ between linear-reverse and cosine-reverse, both trained on the same data and evaluated up to 128K steps. The two schedulers exhibit nearly identical convergence rates across all bins and stages (ratio range 0.97--1.04). This means the mechanistic analysis based on $\eta_r$ \emph{cannot} explain the PPL gap between them (36.5 vs.\ 37.2 in Table~\ref{tab:schedulers}). The source of linear-reverse's advantage likely lies in effects not captured by per-bin convergence rate, such as subtle differences in optimization trajectory or gradient noise structure. We leave a deeper investigation to future work.

\subsection{Convergence-Rate Trajectories Across Training}
\label{app:eta_trajectories}

One might expect low-mask-ratio bins to saturate first, which over a long enough budget would make high-mask-ratio bins the binding constraint and shift the optimal schedule toward forward. Table~\ref{tab:eta_trajectory} reports $\eta_r$ for the lowest and highest mask-ratio bins across the four training stages of the 8B-A1B run in Figure~\ref{fig:mechanistic}. The ratio between them is stable and slightly increasing, with no sign of the convergence toward 1 that saturation would predict. The static-versus-dynamic perplexity gap in Figure~\ref{fig:downstream_1b} behaves consistently, moving from 19.44 versus 19.42 earlier in training to 19.34 versus 19.18, so the separation does not narrow over the last ${\sim}50$B tokens. Both observations are inconsistent with saturation at this scale.

\begin{table}[h]
\centering
\small
\caption{Convergence rate $\eta_r$ for the lowest and highest mask-ratio bins across training stages (8B-A1B, static EC, Nemotron-CC), in the units plotted in Figure~\ref{fig:mechanistic}. The low/high ratio is stable and slightly increasing, so low-mask-ratio bins do not saturate relative to high-mask-ratio ones.}
\label{tab:eta_trajectory}
\begin{tabular}{@{}lccc@{}}
\toprule
Stage (steps) & $\eta_r$, $[0, 0.25)$ & $\eta_r$, $[0.75, 1.0)$ & Low/high \\
\midrule
16K--32K   & 62.2 & 9.8 & 6.35$\times$ \\
32K--64K   & 22.5 & 2.9 & 7.76$\times$ \\
64K--128K  & 7.6  & 1.1 & 6.91$\times$ \\
128K--256K & 2.5  & 0.3 & 8.33$\times$ \\
\bottomrule
\end{tabular}
\end{table}

\section{Scheduler Definitions and FLOPs Equivalence}
\label{app:scheduler_flops}

\subsection{Gaussian Scheduler}

The normalized Gaussian scheduler is defined as:
$$
\tilde{g}(r) = \frac{g(r) - g(0)}{1 - g(0)}, \qquad g(r) = \exp\!\left(-\frac{(r-0.5)^2}{2\sigma^2}\right),
$$
where $g(0) = g(1) = \exp(-1/(8\sigma^2))$. This normalization ensures $\tilde{g}(0)=\tilde{g}(1)=0$ and $\tilde{g}(0.5)=1$, so that the scheduler spans the full $[k_{\min}, k_{\max}]$ range. The Gaussian-reverse scheduler is simply $1 - \tilde{g}(r)$. In our experiments we use $\sigma = 0.22$.

\subsection{FLOPs Equivalence Across Schedulers}

We show that, under uniform masking ratio $r \sim \mathrm{Uniform}(0,1)$, the expected top-$k$ is identical across all schedulers and matches the static baseline. In our experiments, $k_{\min}=8$, $k_{\max}=32$, and the static baseline uses $k=20 = (k_{\min}+k_{\max})/2$.

The per-step FLOPs of the MoE layer are proportional to $k(r)$. The expected computation is:
$$
\mathbb{E}[k(r)] = k_{\min} + (k_{\max} - k_{\min}) \cdot \mathbb{E}[s(r)] = 8 + 24 \cdot \mathbb{E}[s(r)].
$$
It suffices to show $\mathbb{E}[s(r)] = \tfrac{1}{2}$ for each scheduler, which gives $\mathbb{E}[k(r)] = 20$.

\paragraph{Linear / Linear-reverse.}
$\mathbb{E}[r] = \tfrac{1}{2}$ and $\mathbb{E}[1-r] = \tfrac{1}{2}$. Thus $\mathbb{E}[k(r)] = 8 + 24 \times 0.5 = 20.00$. \checkmark

\paragraph{Cosine / Cosine-reverse.}
$\mathbb{E}\!\left[\tfrac{1}{2}(1-\cos\pi r)\right] = \tfrac{1}{2} - \tfrac{1}{2}\int_0^1 \cos(\pi r)\,dr = \tfrac{1}{2} - \tfrac{1}{2}\cdot\!\left[\tfrac{\sin\pi r}{\pi}\right]_0^1 = \tfrac{1}{2}$. Thus $\mathbb{E}[k(r)] = 8 + 24 \times 0.5 = 20.00$. The reverse case follows identically. \checkmark

\paragraph{Gaussian / Gaussian-reverse.}
For the normalized Gaussian $\tilde{g}(r)$, a closed-form expectation is not available, but the symmetry $\tilde{g}(r) = \tilde{g}(1-r)$ constrains the result. Numerical integration with $\sigma=0.22$ gives:
\begin{align*}
\mathbb{E}[\tilde{g}(r)] &= 0.5010, \quad \Rightarrow \quad \mathbb{E}[k_{\text{gau}}(r)] = 8 + 24 \times 0.5010 = 20.02, \\
\mathbb{E}[1-\tilde{g}(r)] &= 0.4990, \quad \Rightarrow \quad \mathbb{E}[k_{\text{gau-rev}}(r)] = 8 + 24 \times 0.4990 = 19.98.
\end{align*}
The deviation from the static baseline ($k=20$) is $\pm 0.02$, or $\pm 0.1\%$ of the expected computation. \checkmark

\paragraph{Summary.}
Table~\ref{tab:flops_equiv} confirms that all schedulers match the static baseline in expected FLOPs.

\begin{table}[h]
\centering
\small
\caption{Expected top-$k$ for each scheduler under $r \sim \mathrm{Uniform}(0,1)$ with $k_{\min}=8$, $k_{\max}=32$.}
\label{tab:flops_equiv}
\begin{tabular}{@{}lrrr@{}}
\toprule
Scheduler & $\mathbb{E}[s(r)]$ & $\mathbb{E}[k(r)]$ & $\Delta$ vs.\ static \\
\midrule
Static ($k=20$)  & 0.5000 & 20.00 & 0.000 \\
Linear           & 0.5000 & 20.00 & 0.000 \\
Linear-reverse   & 0.5000 & 20.00 & 0.000 \\
Cosine           & 0.5000 & 20.00 & 0.000 \\
Cosine-reverse   & 0.5000 & 20.00 & 0.000 \\
Gaussian         & 0.5010 & 20.02 & $+$0.025 \\
Gaussian-reverse & 0.4990 & 19.98 & $-$0.025 \\
\bottomrule
\end{tabular}
\end{table}

\section{Additional SFT Results}
\label{app:sft_samples}

\begin{figure}[h]
  \centering
  \includegraphics[width=\linewidth]{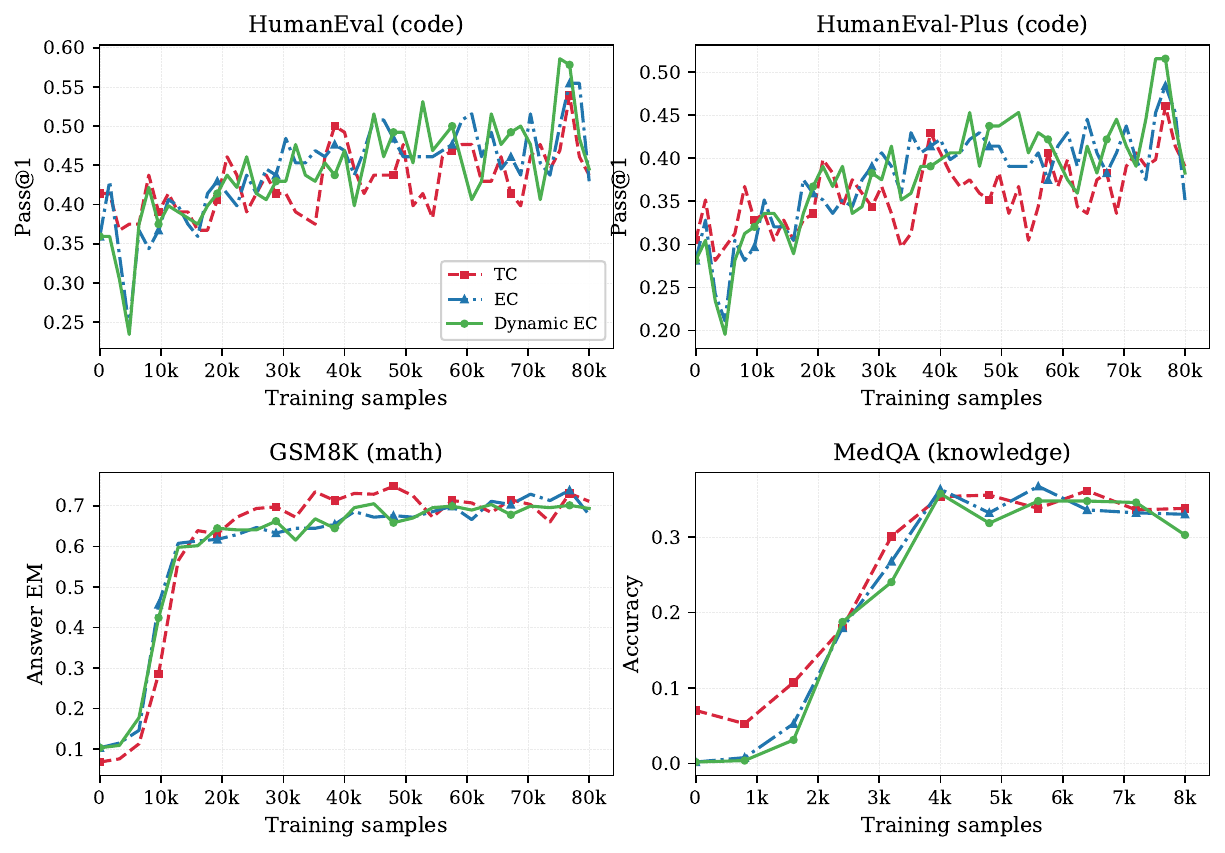}
  \caption{SFT retrofitting results vs.\ number of training samples for TC, EC, and dynamic EC across four benchmarks. Results are consistent with the wall clock time view in Figure~\ref{fig:retrofit_tasks}.}
  \label{fig:retrofit_samples}
\end{figure}

Figure~\ref{fig:retrofit_samples} reports all four SFT benchmarks (HumanEval, HumanEval-Plus, GSM8K, MedQA) plotted against the number of training samples. The trends are consistent with the wall clock time results in Figure~\ref{fig:retrofit_tasks}: EC converges faster and dynamic EC achieves the highest final accuracy.

\begin{figure}[h]
  \centering
  \includegraphics[width=\linewidth]{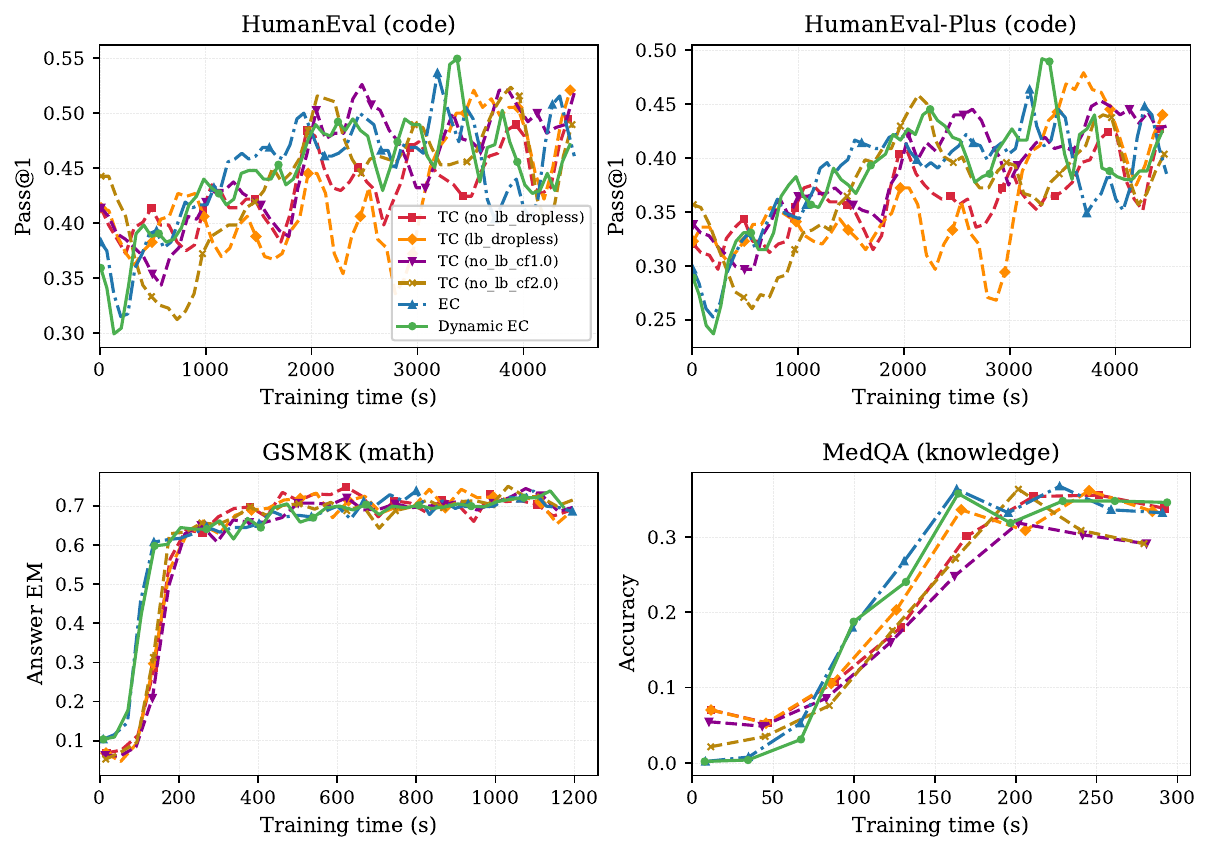}
  \caption{TC routing ablation: comparing four TC variants (no\_lb\_dropless, lb\_dropless, no\_lb\_cf1.0, no\_lb\_cf2.0) against EC and dynamic EC across four benchmarks. All TC variants show similar convergence behavior.}
  \label{fig:tc_ablation}
\end{figure}

Table~\ref{tab:sft_ablation} reports the peak accuracy and evaluation decode time for all TC variants alongside EC and dynamic EC. Across all four TC configurations, the peak accuracy is similar, confirming that the choice of load-balancing strategy within TC has limited impact on final quality. EC and dynamic EC achieve comparable or higher peak accuracy while consistently requiring less decode time, reinforcing the inference speed advantage of expert-choice routing.

\begin{table}[h]
\centering
\small
\caption{Peak accuracy (\%) and evaluation decode time (s) for all TC variants, EC, and dynamic EC across four SFT benchmarks. EC and dynamic EC match or exceed all TC variants in accuracy while decoding faster.}
\label{tab:sft_ablation}
\setlength{\tabcolsep}{3.5pt}
\begin{tabular}{@{}lcccccccccc@{}}
\toprule
& \multicolumn{2}{c}{HumanEval} & \multicolumn{2}{c}{HumanEval+} & \multicolumn{2}{c}{GSM8K} & \multicolumn{2}{c}{MedQA} & \multicolumn{2}{c}{Avg} \\
\cmidrule(lr){2-3} \cmidrule(lr){4-5} \cmidrule(lr){6-7} \cmidrule(lr){8-9} \cmidrule(lr){10-11}
Routing & Pass@1 & Time & Pass@1 & Time & Acc & Time & Acc & Time & Acc & Time \\
\midrule
TC (no\_lb, dropless) & 53.9 & 1369 & 46.1 & 1369 & 74.8 & 672 & 35.5 & 1884 & 52.6 & 1324 \\
TC (lb, dropless)     & 54.7 & 1220 & 50.0 & 1211 & 74.2 & 664 & 36.1 & 1852 & 53.8 & 1237 \\
TC (no\_lb, cf=1.0)   & 54.7 & 1274 & 47.7 & 1261 & 74.4 & 543 & 31.8 & 1645 & 52.2 & 1181 \\
TC (no\_lb, cf=2.0)   & 53.9 & 1288 & 47.7 & 1288 & \textbf{75.0} & 647 & 36.3 & 1830 & 53.2 & 1263 \\
\midrule
EC                    & 55.5 & \textbf{1008} & 48.4 & \textbf{1008} & 73.8 & \textbf{459} & \textbf{36.7} & \textbf{1371} & 53.6 & \textbf{962} \\
Dynamic EC            & \textbf{58.6} & 1056 & \textbf{51.6} & 1056 & 73.8 & 461 & 35.7 & 1378 & \textbf{54.9} & 988 \\
\bottomrule
\end{tabular}
\end{table}

\section{Additional Related Work}
\label{app:additional_related_work}

\paragraph{Adaptive computation.}
Prior work adapts computation along the \emph{depth} dimension: early exit and confidence-based halting~\citep{schuster2022confident,elhoushi2024layerskip}, Mixture-of-Depths routing~\citep{raposo2024mixture,bae2025mixture}, and looped/recursive transformers~\citep{lan2020albert,hutchins2022block,fan2025looped,bae2025relaxed,geiping2025scaling}. Our approach is orthogonal: we adapt computation along the \emph{timestep} dimension, which is unique to iterative generative models.

\paragraph{Timestep-adaptive computation in diffusion models.}
A growing body of work shows that different diffusion timesteps have fundamentally different learning dynamics. In continuous diffusion, process increment analysis~\citep{kim2025speed} and curriculum-based difficulty measurement~\citep{kim2025curriculum} reveal that low-noise timesteps are harder to learn, while Min-SNR weighting~\citep{hang2023efficient} addresses cross-timestep gradient conflicts and noise schedule analysis~\citep{lin2024common} identifies undertrained timestep regions. In the discrete setting, DiffusionBERT~\citep{he2023diffusionbert} shows that token-level masking order affects generation quality, and MDLM~\citep{sahoo2024simple} demonstrates that timestep sampling strategies significantly impact training variance. At the model level, AdaDiff~\citep{tang2024adadiff} and DyDiT~\citep{zhao2025dynamic} dynamically adjust model width and depth per timestep, while MoE-based approaches target vision diffusion transformers: EC-DIT~\citep{sun2025ecdit} shows that EC routing implicitly adapts to timesteps, DiffMoE~\citep{shi2025diffmoe} learns a capacity predictor for per-timestep allocation, and Diff-MoE~\citep{cheng2025diffmoe} injects timestep conditioning into expert routing. All of these prior methods target continuous diffusion for vision. Our work is the first to study timestep-adaptive expert capacity in \emph{discrete masked diffusion for language}, with explicit capacity scheduling and mechanistic analysis of why low-mask-ratio steps benefit most from additional computation.

\paragraph{Parallel decoding and inference for diffusion LLMs.}
A complementary line of work accelerates DLM inference through parallel decoding strategies. Fast-dLLM~\citep{wu2025fastdllm} introduces KV caching and confidence-aware parallel decoding for DLMs, while Fast-dLLM~v2~\citep{wu2025fastdllmv2} and Esoteric LMs~\citep{sahoo2026esoteric} bridge autoregressive and diffusion paradigms to enable efficient block-wise generation. Several methods focus on improving parallel decode quality: dParallel~\citep{chen2026dparallel} uses certainty-forcing distillation, Hierarchy Decoding~\citep{qi2026hierarchy} applies a divide-and-conquer strategy, adaptive parallel decoding~\citep{israel2025adaptive} dynamically adjusts decode width, and FreeDave~\citep{wu2026freedave} achieves lossless parallel decoding via integrated draft-and-verification. ParallelBench~\citep{kang2025parallelbench} provides a systematic benchmark revealing that parallel decoding quality depends heavily on token dependencies. On the sampling side, path planning methods~\citep{peng2026pathplanning,peng2026planneraware} optimize the denoising trajectory, while corrective approaches~\citep{zhang2026corrective,kim2025prism} enable DLMs to detect and revise erroneous tokens during iterative refinement. Our work is orthogonal to these inference-time methods: we optimize the \emph{training-time} computation allocation via expert capacity scheduling, which could be combined with parallel decoding at inference.

\end{document}